\documentclass[sn-mathphys,Numbered]{sn-jnl}

\usepackage{graphicx}%
\usepackage{multirow}%
\usepackage{amsmath,amssymb,amsfonts}%
\usepackage{amsthm}%
\usepackage{mathrsfs}%
\usepackage[title]{appendix}%
\usepackage{xcolor}%
\usepackage{textcomp}%
\usepackage{manyfoot}%
\usepackage{booktabs}%
\usepackage{algorithm}%
\usepackage{algorithmicx}%
\usepackage{algpseudocode}%
\usepackage{listings}%

\begin{document}

\title[Article Title]{Transformers and Large Language Models for Chemistry and Drug Discovery}


\author*[1,2]{\fnm{Andres} \sur{M Bran}}\email{andres.marulandabran@epfl.ch}

\author*[1,2]{\fnm{Philippe} \sur{Schwaller}}\email{philippe.schwaller@epfl.ch}

\affil[1]{\orgdiv{ISIC, EPFL}, \orgname{Laboratory of Artificial Chemical Intelligence (LIAC)}, \orgaddress{\city{Lausanne}, \postcode{1050}, \state{VD}, \country{Switzerland}}}

\affil[2]{\orgdiv{EPFL}, \orgname{National Centre of Competence in Research (NCCR) Catalysis}, \orgaddress{\city{Lausanne}, \postcode{1050}, \state{VD}, \country{Switzerland}}}


\abstract{Language modeling has seen impressive progress over the last years, mainly prompted by the invention of the Transformer architecture, sparking a revolution in many fields of machine learning, with breakthroughs in chemistry and biology. In this chapter, we explore how analogies between chemical and natural language have inspired the use of Transformers to tackle important bottlenecks in the drug discovery process, such as retrosynthetic planning and chemical space exploration. The revolution started with models able to perform particular tasks with a single type of data, like linearised molecular graphs, which then evolved to include other types of data, like spectra from analytical instruments, synthesis actions, and human language. A new trend leverages recent developments in large language models, giving rise to a wave of models capable of solving generic tasks in chemistry, all facilitated by the flexibility of natural language. As we continue to explore and harness these capabilities, we can look forward to a future where machine learning plays an even more integral role in accelerating scientific discovery.}


\maketitle

\section{Introduction}\label{sec1}

The capacity to process and accurately model human language has been a persistent pursuit within the machine learning community \cite{bahdanau_neural_2016, cho_learning_2014, hochreiter_long_1997, sutskever_sequence_2014, vaswani_attention_2017}. The belief is that language is intrinsic to human reasoning capabilities, thus successful language modeling could open the door to numerous applications, enhancing various information processing tasks with the potential to revolutionize several industries \cite{bommasani_opportunities_2022, bubeck_sparks_2023}. The field of natural language processing has witnessed significant advancements in recent years, thanks to improved computing infrastructure, breakthroughs in algorithms, and the proliferation of abundant and accessible data \cite{gao_pile_2020}. Language and technical terminology also play a pivotal role in the domain of chemistry, which serves as the fundamental basis for drug discovery and development. Analogous to human language, understanding and accurately modeling the language of chemistry is crucial for effective research and development in the pharmaceutical industry. By applying the advancements in language modeling and processing from the machine learning community to the domain of chemistry, it is possible to facilitate drug discovery by efficiently analyzing and interpreting vast amounts of chemical data and literature.

Introduced in 2017, the Transformer architecture revolutionized natural language processing \cite{vaswani_attention_2017}. The Transformer is a type of neural network initially developed for neural machine translation. In its original architecture, the Transformer consists of an encoder, which encodes a sentence in the source language (e.g., French) and a decoder, which word-by-word decodes the translated sentence in the target language (e.g. English). The power of the Transformer model comes from its core building blocks, so-called attention layers \cite{bahdanau_neural_2016}. Those attention layers are excellent at capturing the meaning of words and subwords in their context. For this to work, text is split into subwords and inputted into the Transformer, which encodes the sequence of subwords and learns to use its attention layers to connect relevant pieces of information. The Transformer then generates another sequence of tokens as output, using its decoder part. 

The original Transformer and variants of it, such as encoder-based and decoder-based architectures, have shown remarkable results in a range of language modeling tasks, including translation \cite{raffel_exploring_2020}, sentiment classification \cite{devlin_bert_2019, reimers_sentence-bert_2019}, and summarization \cite{liu_generating_2018, kryscinski_neural_2019}, to name a few. Over time, the architecture of these models has evolved, with the most popular variations being decoder-only structures, which make up the foundation of many contemporary, large language models (LLMs), such as ChatGPT and GPT-4 \cite{openai_gpt-4_2023}. 

\begin{figure}[ht!]
    \centering
    \includegraphics[width=\linewidth]{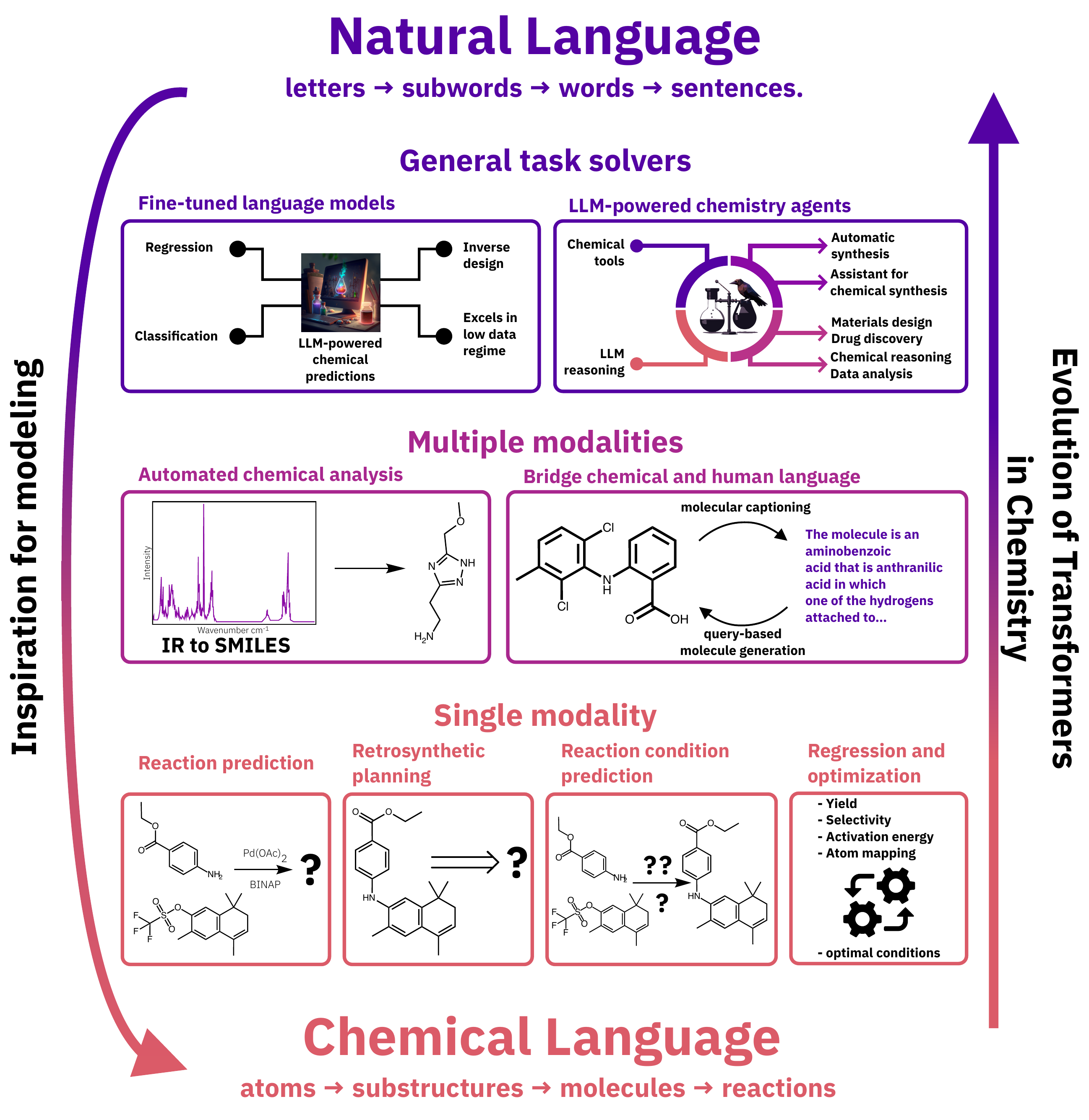}
    \caption{Advances in Natural Language Processing have inspired applications in Chemistry. Over time, the gap between chemical Language and natural Language is being closed by including additional modalities. Most recent works present general task solvers for chemistry, capable of chemical reasoning and automatic synthesis, among others.}
    \label{fig:fig_1}
\end{figure}

Similar to human language, where the tokens — the subunits in which the text is broken down — are words and subwords, we can imagine splitting proteins and peptides into sequences of amino acids, or molecules and chemical reactions into sequences of atoms (Figure \ref{fig:fig_1}). Such analogies between human language and (bio)chemistry extended the influence of Transformers in fields far beyond the boundaries of natural language processing impacting numerous scientific disciplines.

A prime example of Transformers' applications is their pivotal role in AlphaFold2, a leading system for predicting the 3D structure of proteins \cite{jumper_highly_2021}. The emergence of AlphaFold2 not only addressed a long-standing challenge in biochemistry but also catalyzed a surge of research, thereby becoming one of the most prevalent tools in the biochemistry community \cite{yang_alphafold2_2023,akdel_structural_2022}.

Chemistry has also benefited from these advancements. Inspired by developments in other areas, researchers have adapted a number of chemical tasks in the form of text sequences (Figure \ref{fig:fig_1}). This, together with the rise of open datasets and benchmarks \cite{huang_therapeutics_2021,kearnes_open_2021, lowe_extraction_2012, wu_moleculenet_2018}, sparked a revolution that started with clearly defined chemical challenges fundamental to the drug development process, like reaction outcome prediction and retrosynthetic planning. Models at this stage are trained with only molecules as both input and output, making them single-modal models.

The revolution then continued in a new direction, with attempts to include additional types of data like spectra from analytical techniques, and human language as in synthesis procedures, giving rise to a wave of chemical multimodal models. The current stage aims to definitively bridge the gap between the modeling of chemical and natural language (Figure \ref{fig:fig_1}). With applications based on large language models, researchers have demonstrated how natural language can serve as interfaces for chemical reasoning and task solving across fields.

We are now in an era marked by significant progress in a host of tasks that are fundamental to the drug development process, with systems that go beyond single task solving and are capable of implementing complete pipelines in this process, accelerating chemical research. The influence of these Transformer models in the realm of chemistry is thus profound, highlighting their pivotal role in shaping the future of chemistry and drug discovery. 

The next sections will briefly introduce textual representations of molecules and reactions, and then continue to discuss task-specific Transformers for single-modality and multi-modality tasks. Lastly, we will address large language models and potential uses in chemistry and drug discovery.

\section{Modeling the language of organic chemistry}

Chemistry, in many respects, resembles a language \cite{cadeddu_organic_2014}. Not only is most of the information in chemistry conveyed through human language, but the rules governing chemical transformations also form a distinct language themselves. Accurately modeling this language advances our understanding of its rules, opening up applications such as automatic retrosynthetic planning and efficient chemical space exploration \cite{wolos_computer-designed_2022}, while also shedding light on the intricate grammar of organic chemistry \cite{schwaller_extraction_2021}.

However, the chemical language is not a conventional language like English or Mandarin, which operate on text. In organic chemistry, grammar operates on a spectrum of molecular graphs and reaction conditions, making the direct application of Transformers to chemistry a challenge. Overcoming this obstacle is thus vital and, as we will explore, can be accomplished using linear string molecular representations that have been in use for decades, with new revisions and proposals in recent years \cite{weininger_smiles_1988, oboyle_deepsmiles_2018, krenn_self-referencing_2020, krenn_selfies_2022, heller_inchi_2015, brammer_tucan_2022}

\subsection{Text Representations of Molecules and Reactions}

Chemistry has long been acknowledged as a fundamentally inventorial science, wherein new molecules and reactions are discovered, analyzed, and to a certain extent cataloged in databases \cite{restrepo_chemical_2022}. Researchers rely not only on scientific articles and patents but also on resources like handbooks and, more recently, computational databases, to advance their work. To streamline the storage and querying of these data, simplified molecular-input line-entry system (SMILES) strings have been proposed and utilized since the 1980s \cite{weininger_smiles_1988}.

\begin{figure}[ht!]
    \centering
    \includegraphics[width=\linewidth]{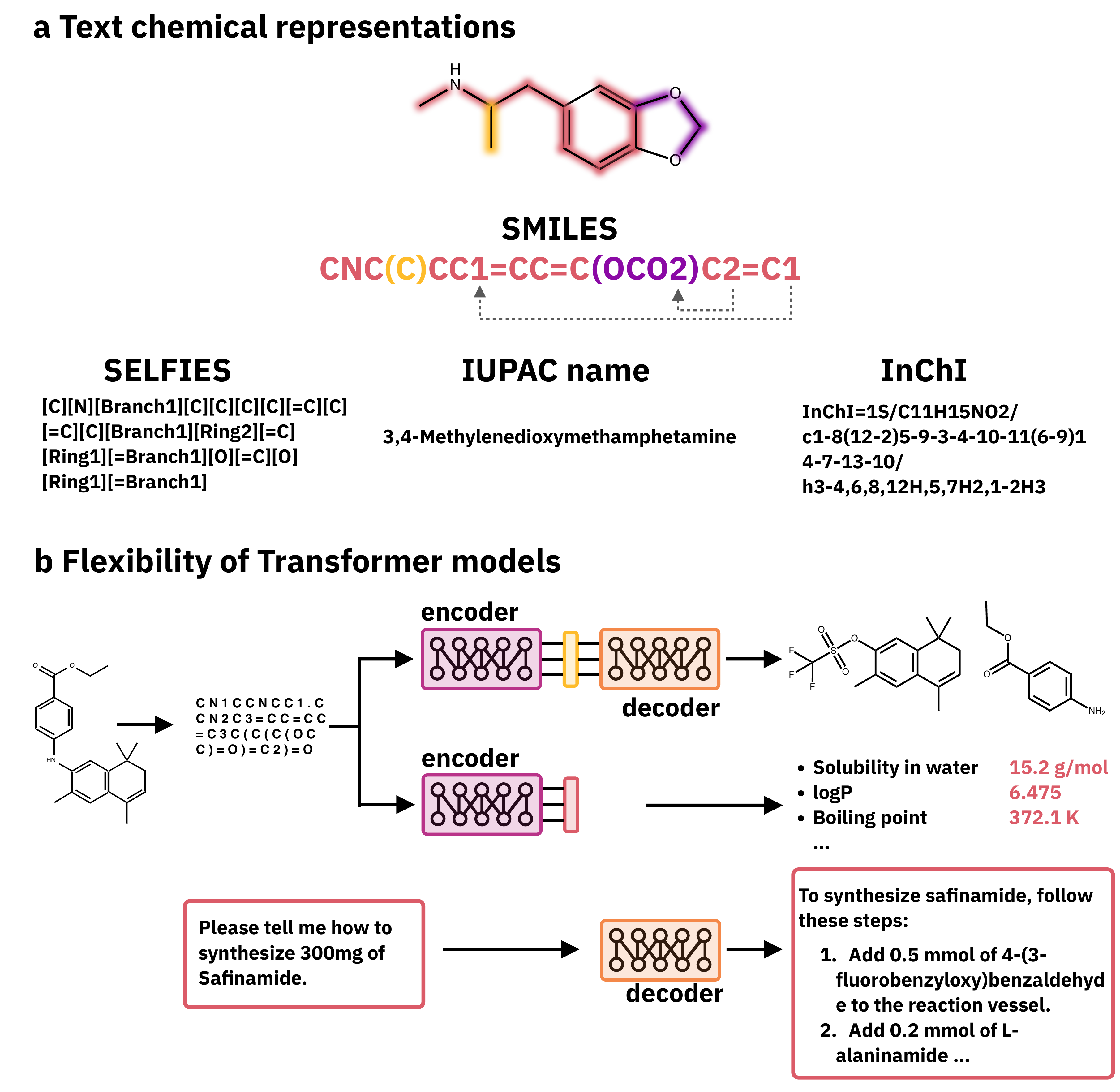}
    \caption{The language of organic chemistry operates on molecular graphs, which can easily be converted into multiple text formats (a), facilitating the use of Transformers in their multiple forms, for a variety of tasks in Chemistry (b). }
    \label{fig:fig_2}
\end{figure}

The process of linearizing molecular graphs involves choosing an atom and sequentially enumerating all other atoms in the molecule from this point (Figure \ref{fig:fig_2}a). Special characters are assigned to specify bond types, branches, rings, stereochemistry, and other pertinent information for molecular representation. SMILES leverage the fact that molecular graphs adhere to certain chemical rules, such as different types of atoms needing to respect their valence. This makes it feasible to represent a large portion of organic chemistry as sequences of characters.

As molecular machine learning applications have emerged, the limitations of these representations —such as their lack of robustness— have become increasingly apparent. In generative models, this flaw can lead to the production of entirely invalid molecules, becoming an additional hurdle in molecular discovery applications \cite{kusner_grammar_2017, gomez-bombarelli_automatic_2018}. To mitigate this issue, researchers have introduced Self-Referencing Embedded Strings (SELFIES), a novel string-based representation that effectively addresses the robustness concern \cite{krenn_selfies_2022}. This new representation has found numerous applications, particularly in drug discovery and molecular generation, due to its unique formulation that guarantees the correct mapping of any such string to a valid molecule. While other representations have been proposed in the past to standardize the language (IUPAC names, InChI \cite{heller_inchi_2015}), or to address the limitations of current representations for deep learning applications (DeepSMILES \cite{oboyle_deepsmiles_2018}), we will not delve further into these representations in this discussion \cite{krenn_selfies_2022, ozturk_exploring_2020}.

With these molecular text representations in hand, chemical reactions can easily be encoded by leveraging the syntax of chemical equations. In that sense, sets of molecules —e.g. the set of reactants— are represented by listing all molecules with a dot ``.'' between them. Reactants are separated from the products using the symbol ``$>$'', representing the arrow, and other details like catalysts and reagents are placed in between the ``$>>$'' symbol. The resulting reaction representation has the form ``A.B$>$catatlyst.reagent$>$C.D'', where A and B are reactants and C and D products. This notation is commonly used in SMILES and called a ``reaction SMILES''.

\subsection{Task-specific Transformers}
The conversion of chemical problems into sequences of tokens —in the form of a language— has unlocked the transformative potential of the Transformer architecture within the chemistry domain. This shift has led to remarkable model performances in prediction tasks in chemistry, such as retro- and forward-synthesis \cite{pesciullesi_transfer_2020, schwaller_molecular_2019, schwaller_predicting_2020}, as well as molecular regression \cite{li_mol-bert_2021, chithrananda_chemberta_2020, ahmad_chemberta-2_2022} and reaction classification \cite{schwaller_mapping_2021}. Moreover, it has set the stage for other, more diverse tasks, like inferring experimental procedures \cite{vaucher_automated_2020, vaucher_inferring_2021}, that go beyond operations on molecular graphs and require a deeper understanding of experimental conditions and standard procedures, a feat only achievable through modeling human language.

Achieving success in this wide range of tasks has been facilitated by different variations of the Transformer architecture. Depending on the application, these variations leverage different parts of the architecture (Figure \ref{fig:fig_2}b), leading to encoder-decoder models —useful for tasks involving conversion of a sequence into another, like translation—, encoder-only —to extract rich representations from data—, and decoder-only architectures, mostly used in generative applications (Figure \ref{fig:fig_2}b).

\subsubsection{Chemical translation tasks}
The primary motivation for the formulation of the Transformer architecture by \citet{vaswani_attention_2017} was translation. In this task, input text in one language is converted into corresponding text in another language, logically leading to an architecture with an encoder part linked to a decoder (Figure \ref{fig:fig_2}b).

Steps to exploit this design were taken by \citet{schwaller_molecular_2019}, they introduced the Molecular Transformer and treated the task of reaction prediction as a translation task. In this perspective, a Transformer model learns to “translate” from precursors to products SMILES. The approach proved highly successful, establishing itself as one of the state-of-the-art methods for this task \cite{tetko_state---art_2020}. Subsequent attempts applied a similar approach to other tasks, like retrosynthetic planning \cite{schwaller_predicting_2020}, where a model learns to predict the reactants (and reagents) necessary to produce a given product. The model is then successively applied to the predicted reactants iteratively, to construct a retrosynthetic tree that maps to commercially available building blocks. Recent work used similar retrosynthesis models to tackle the challenge of suggesting diverse candidate reactions \cite{toniato_enhancing_2023}, as well as prompting, steering and unbiasing the proposed disconnections \cite{thakkar_unbiasing_2023}.

The Molecular Transformer approach was further extended by \citet{irwin_chemformer_2022}, who proposed the Chemformer, a model pre-trained on a variety of chemical tasks, that can then be specialized for specific applications. This methodology achieved even greater success while offering additional flexibility and transferability to new tasks. 

An interesting variation of the architecture came from an attempt to directly encode molecules as molecular graphs, for translation tasks like reaction outcome prediction. \citet{tu_permutation_2021} used a custom-designed graph encoder, with a Transformer decoder, trained to translate molecular graphs to SMILES. This approach proved valuable in forward- and retrosynthesis tasks, improving over other Transformer-based baselines, due to the permutational invariance of the novel graph-encoder.

\subsubsection{Unsupervised learning and feature extraction.}

Transformers have also been applied in various fields to generate rich vectorial representations of text that encapsulate context and general features \cite{reimers_sentence-bert_2019, mikolov_distributed_2013}. Such methods have been used for sentiment analysis, where a given text is classified based on the sentiment it conveys, a useful tool for assessing customer satisfaction with a specific product. The process of encoding this text as vectors is known as representation learning. For this task, only the encoder part of the Transformer architecture is used, as there is no need for text generation.

In the realm of chemistry, representation learning is a critical issue \cite{duvenaud_convolutional_2015}. It enables the transformation of molecules and reactions into vector representations, thus enabling a multitude of downstream applications. These include similarity assessment for database lookups, as well as regression and classification for property prediction tasks, such as predicting reaction yields or identifying toxic compounds — both of which are crucial to the drug development process.

Adopting similar techniques, \citet{wang_smiles-bert_2019} trained a model to generate reaction representations, and compared them against commonly used hand-engineered molecular representations, showing gains in accuracy in a number of downstream regression tasks, demonstrating the power of Transformer encoders for tasks in chemistry.

In another study, the decoder side of the Transformer was replaced by a classification layer, and the model was trained to predict the class of chemical reactions \cite{schwaller_mapping_2021}. The resulting vector representations, in addition to achieving high classification accuracies, were used for the visualization and exploration of a database of chemical reactions from patents. This revealed patterns in the data that grouped reactions by class, but also by data source, and relevant product properties, like logP, number of H donors, and others \cite{schwaller_machine_2022}. This study showcased how such numerical representations can encode intricate details of chemical reactions, thereby facilitating effective clustering and classification of reactions. Importantly, this was achieved without the need for hand-crafted molecular or reaction features, demonstrating the model's capacity to learn purely from extensive reaction databases. In a similar vein, a regression layer, instead of a decoder layer, was used for predicting reaction yields, achieving outstanding results on datasets from high-throughput-experimentation platforms \cite{schwaller_prediction_2021, neves_global_2023}. 

Other works have further advanced the applications of encoder-only Transformers, with enhancements in training procedures \cite{li_mol-bert_2021, chithrananda_chemberta_2020, ahmad_chemberta-2_2022}, modifications in the architecture \cite{ross_large-scale_2022}, and applications in larger machine learning systems towards the solution of more specific objectives \cite{wu_molformer_2023}.

Similar ideas have also been implemented, with a focus on biochemistry. \citet{rives_biological_2021} trained a transformer model on 250 million unlabeled protein sequences, with the goal of learning the “language of proteins”, an approach known as unsupervised learning. The resulting representations not only enabled state-of-the-art property predictions for proteins, but also predictions of variant effects and protein folding \cite{lin_evolutionary-scale_2023}. The model was further shown to generalize beyond naturally occurring proteins \cite{verkuil_language_2022}, paving the way for applications in de novo generation of proteins for specific applications.

Perhaps one of the most intriguing applications of Transformers in chemical representation learning, also leveraging unsupervised learning, emerged from an attempt to interpret their inner workings. Through a detailed examination of the attention weights in a Transformer model trained on unlabeled chemical reactions, \citet{schwaller_extraction_2021} discovered that  these models create internal representations that connect atoms from the reactants to other atoms in the products. This almost serendipitous discovery led to the creation of RXNMapper, an open-source algorithm that accurately calculates the atom mapping in a given chemical reaction. RXNMapper has been shown to outperform other approaches —many of which were under closed licenses— in terms of speed, parallelizability, and accuracy. A similar approach was recently investigated on enzymatic reactions, where the attention weights could be used to identify active sites in protein sequences \cite{teukam_language_nodate}.

\subsubsection{Articulating chemical language and other modalities}

Chemical transformations are not confined solely to the realm of chemical structures. They are enhanced and accessed through a variety of other data types, or modalities. These include human language used for describing molecules and experimental results, as well as the experimental results themselves, which may be presented in formats such as numerical arrays or images, among others.

Considering this, scientists have proposed tasks designed to bridge the divide between the language of molecules and human language. One such task is known as molecular captioning, which involves describing a specific molecule in natural language \cite{edwards_translation_2022}. The description can cover a wide range of features such as molecular scaffolds, the source of the substance, drug interactions, and more, all expressed in simple English.

Some researchers have expanded on this concept \cite{christofidellis_unifying_2023, schwaller_predicting_2020}, enabling a smooth interconversion from molecules to natural language and vice versa. This advancement has given rise to versatile models capable of not just molecular captioning, but also generating molecules in response to text queries, or carrying out molecule-to-molecule conversion tasks, like predicting reaction outcomes and retrosynthesis \cite{christofidellis_unifying_2023}. While these models are still in their early stages of development, they show immense potential for the future of drug discovery as they continue to be developed and refined.

Another task, highly relevant for the design of automatic robotic platforms for synthesis, is the prediction of experimental steps. A step missing from retrosynthetic planning is the experimental realization, which involves steps such as adding substances, stirring, and purifying, details not contained in predicted reactions. To bridge this gap, \citet{vaucher_automated_2020, vaucher_inferring_2021} compiled a database of reactions with associated action sequences, and trained models to generate actionable sequences of steps from a given reaction in SMILES format. 
Other tasks aim to link experimental results with molecular structures, effectively addressing the challenge of structural elucidation. A pioneering attempt using Transformers was recently made by \citet{alberts_leveraging_2023}, who created a database of molecules and computationally generated IR spectra to train a Transformer model for this purpose. IR spectra, often overlooked for this task due to their complexity, were represented as a sequence of numerical values encoded in text alongside the chemical formula. These were then fed to the model, which was tasked with predicting the SMILES string of the underlying molecular structure. This method achieved a 45\% accuracy rate in structure prediction, while also surpassing all previous efforts in predicting functional groups from IR spectra. This illustrates how underutilized data, such as IR spectra, can be harnessed for tasks as intricate as structural elucidation, despite their initial difficulty in interpretation.

These excellent applications of Transformers across a range of tasks in chemistry underscore just a fraction of the potential these models possess, as well as the wealth of latent knowledge embedded in chemical databases. So far, our discussion has focused on task-specific applications, where a predefined task is established and models are specifically trained to solve that task. One of the advantages of language is its flexibility, enabling it to express various tasks in similar forms of the same language. The upcoming section will delve into the potential of more generalized language models for chemistry.

\section{Advanced applications: Beyond task-specific models}

The last few years have seen a remarkable rise in the power and popularity of foundation models: models pre-trained on vast amounts of data that can easily be adapted to other, more specific tasks \cite{bommasani_opportunities_2022, openai_gpt-4_2023}. Such pre-training is conducted using rich databases of human text extracted from the internet, among others, to make the model learn the language along with general knowledge about the world. This rise has been achieved mainly through the escalation of Transformer architectures to large computational budgets and vast databases, leading to models capable of producing text that matches human-level quality in a wide range of scenarios \cite{bommasani_opportunities_2022, openai_gpt-4_2023}.

The process of fine-tuning these models leverages their pre-learned language abilities while tailoring them toward a specific objective where less data might be available \cite{raschka_finetuning_2023, zhang_llama-adapter_2023}. A prime example of this is ChatGPT, a large language model fine-tuned for conversation. Its release has not only catapulted these models into popularity but has also reignited profound philosophical discussions about the nature of intelligence \cite{bubeck_sparks_2023}. However, it has also raised serious concerns about potential issues such as the spread of misinformation. The power and accessibility of ChatGPT have set the world on a new trajectory, prompting us to rethink how we produce and consume media, while also highlighting the need for careful consideration of the potential implications.
The success and popularity of ChatGPT can be attributed to two key factors. Firstly, its freely available and highly user-friendly chat interface, which makes interaction with the model straightforward, and secondly, its impressive capabilities, which often extend beyond the tasks the model was initially trained for. These capabilities not only demonstrate the power of ChatGPT and models of its kind but also hint at their potential for innovative applications.

\begin{figure}[ht!]
    \centering
    \includegraphics[width=\linewidth]{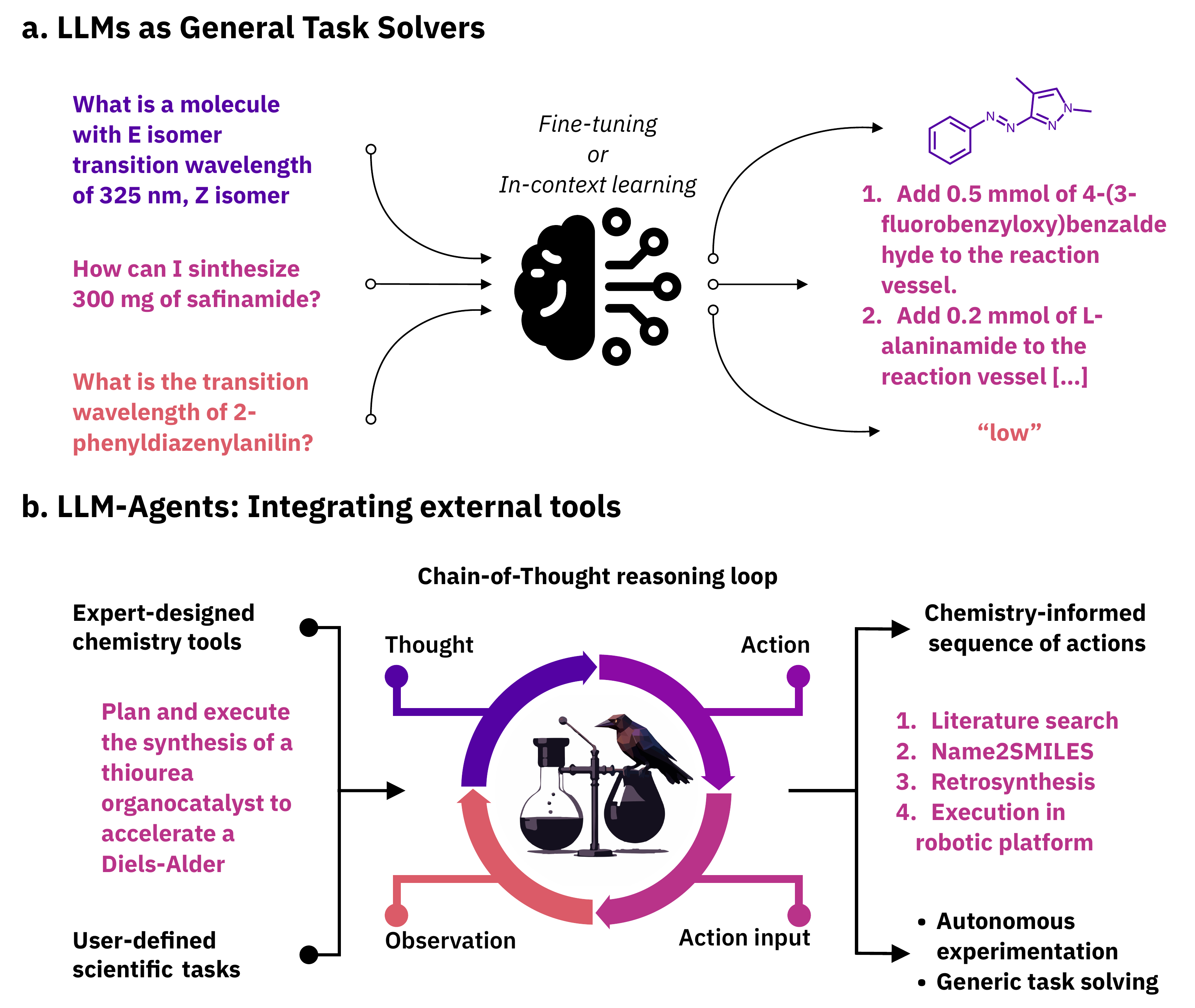}
    \caption{Recent advances in Large Language Models have launched a new era of Transformers in chemistry. Their capabilities allow for a. general task solvers thanks to their flexibility and knowledge transferability \cite{jablonka_is_2023, ramos_bayesian_2023} and b. agent architectures, capable of integrating virtually unlimited modalities, in the form of computational tools \cite{jablonka_14_2023, bran_chemcrow_2023, boiko_emergent_2023}.}
    \label{fig:fig_3}
\end{figure}

\subsection{On the capabilities of Large Language Models}
The rise of data-hungry machine learning algorithms, coupled with the increasing availability of data, has set a trend for scaling models to the maximum sizes that hardware constraints allow. As these models increase in size, they correspondingly improve in their capacity to perform the tasks they were originally trained for. This trend is particularly predictable in the realm of language models, and it manifests in the form of what researchers refer to as scaling laws \cite{hoffmann_training_2022}. 

These scaling laws serve as a valuable tool for researchers, enabling them to identify performance trends and make accurate predictions about the capabilities of significantly larger models. However, the process of scaling does not simply enhance existing capabilities. As these models grow, they reach certain critical thresholds where not only their existing capabilities are enhanced, but entirely new capabilities are observed.

These new abilities are collectively referred to as emergent capabilities \cite{wei_emergent_2022}. They represent a fascinating aspect of model scaling, as they are not present or even predictable in smaller models, but suddenly appear as the models increase in size. These emergent capabilities offer exciting potential for the future of machine learning and its application in various fields, including chemistry.

\citet{wei_emergent_2022} demonstrated how language models below certain computing budgets display somewhat random behavior across a range of tasks. However, once a certain model size is reached, sudden and significant improvements in performance are observed. Other capabilities are observed in the form of augmented prompting strategies, such as Chain of Thought (CoT) reasoning \cite{wei_chain--thought_2023}. In this approach, models are instructed to solve a task by following a step by step reasoning sequence. Another emergent capability is instruction following \cite{ouyang_training_2022}, where LMs are given a task in the form of a set of instructions to follow.

Interestingly, these techniques generally have a negative impact on the performance of smaller models \cite{wei_chain--thought_2023}. However a positive effect on performance is observed once models reach certain threshold sizes \cite{wei_emergent_2022}.

Emergent capabilities, therefore, enable language models to effectively tackle a variety of tasks that involve reasoning. This is achieved without explicit training and with the flexibility of a text query in natural language. Considering the remarkable capabilities and transformative potential of large language models in various fields, it naturally leads us to question what benefits these models could bring to the field of chemistry. This will be the focus of our exploration in the following sections.

\subsection{Large Language Models in chemistry}
Much of this chapter has been devoted to the application of the Transformer architecture to ingeniously-encoded specific chemical tasks. However, it is  important to note that a significant portion of information in chemistry is conveyed through human language, which includes explanations of reaction mechanisms, descriptions of modes of action of drugs, to name a few. Reasoning in chemistry is thus fundamentally articulated in human language, even though it is complemented by other non-text objects like graphs and images. This observation raises the question of whether Large Language Models (LLMs) can replicate this level of chemical reasoning and, if so, to what extent.

\subsubsection{Tailoring techniques: fine-tuning and in-context learning}
One of the main goals of developing general, large pretrained language models, is being able to further tune them for specific applications. Techniques for this include fine-tuning, where a model’s parameters are further optimized for a new task. This technique has been the dominating paradigm in machine learning for years, with excellent results in a number of applications \cite{wei_finetuned_2021, yin_fine-tuning_2017, howard_universal_2018}. 

A new paradigm has also become the target of investigations with the rise of LLMs, so-called in-context learning. This feature can be activated simply by providing the model with a task description, along with a small set of examples that serve as “training data”. This method proves particularly useful when data is scarce or obtaining it is costly or time-consuming, as is often the case in chemistry laboratories. LLMs have demonstrated impressive performance across a range of tasks using this method, with their strong performance believed to stem from knowledge transfer across different pre-training sources.

In line with this, \citet{jablonka_is_2023} demonstrated that LLMs like GPT-3 can effectively solve various tasks in chemistry and materials science by fine-tuning LLMs, to a good level of approximation. The authors chose a range of tasks for which datasets are typically limited, such as predicting the transition wavelength of molecules, the phase of solid solutions in high entropy alloys, or even inverse design. Given the limited datasets available for these tasks, machine learning solutions have traditionally relied on heavily engineered and specialized algorithms \cite{dai_retrosynthesis_2020, duvenaud_convolutional_2015, zhang_chemistry-informed_2022}. These algorithms aim to directly incorporate chemical knowledge into the model architectures. In parallel, extensive research has been devoted to molecular and reaction representations, with some efforts also attempting to bias the representations using prior chemical knowledge to better encode relevant features in data \cite{jorner_kjelljornermorfeus_2022}. All these endeavors aim to make the most of existing knowledge, enabling models to learn as much as possible from the few available data points.

Interestingly, \citet{jablonka_is_2023} demonstrated that fine-tuning and in-context learning can perform on par with, and in some instances even outperform, these specialized techniques, particularly when data is limited. The high performance of in-context learning, combined with its ease of use and flexibility, makes it one of the most impressive applications of LLMs in chemistry to date. This technique holds the potential to revolutionize the way machine learning is utilized in the scientific field, by rapidly highlighting complex correlations in data.

Another application of regression, which holds significant interest in chemistry and drug discovery cycles, is optimization. This process involves modifying an object until a property of interest reaches a desired value. Typically, this requires a large number of measurements of the desired property, which can be quite costly in chemistry use-cases. Applications of this include yield/selectivity optimization in chemical reactions and the generation of molecular candidates with target properties. Bayesian Optimization (BO) has recently been proposed as a solution to such problems in chemistry \cite{rankovic_bayesian_2023, shields_bayesian_2021}, particularly in situations where data is small. However, BO requires uncertainty-calibrated regression methods, which sets it apart from conventional regression. 

In line with the concept of in-context learning, \citet{ramos_bayesian_2023} proposed a system that utilizes GPT models to perform regression while also incorporating uncertainty. This approach enables BO without the need for any feature engineering or fine-tuning. The flexibility of this method allowed the team to perform catalyst and molecular optimization using only the synthesis procedure of the catalyst as input. This work represents a paradigm shift in drug discovery and molecular design. For the first time, it showcases a direct map from synthesis procedure into property space, effectively overcoming issues like the synthesizability of proposed molecules, a key limitation of structure-based generative models.

\subsubsection{Molecular generation}
Another fascinating application of the generative capabilities of language models is molecular generation. This area, which is of significant importance in the drug discovery process, has been largely dominated by models that generate molecules in the form of linear string representations, such as SMILES or SELFIES \cite{wang_transformer-based_2022, rothchild_c5t5_2021, bagal_molgpt_2022}. While this approach has proven successful due to the ease of training these models \cite{bengio_flow_2021, born_regression_2023, jablonka_is_2023}, its successful application is contingent on the ability to specify substances as graphs and their subsequent conversion to a linear string representation. However, this approach is only suitable for a subset of organic molecules. Other substances, such as macromolecules and materials, necessitate more comprehensive representations. A complete and accurate representation of these substances can only be achieved by specifying atomic positions, boundary conditions, and other factors. This requirement presents a significant challenge and limitation to the current methods of molecular generation. To address these limitations, \citet{flam-shepherd_language_2023} proposed using language models for structure generation, directly generating them with three-dimensional atomic positions. Besides being innovative and valid, the generated structures can be obtained by training models in a variety of formats used for crystals, proteins, and more. This work also demonstrates performance comparable to expert-designed, state-of-the-art algorithms for molecular generation based on graphs, while overcoming the limitations mentioned earlier.

\subsection{Language Model-powered Agents}
Among the most useful emergent abilities of language models are step-by-step reasoning, activated through chain-of-thought (CoT) prompting, and their capacity to effectively use tools \cite{schick_toolformer_2023}. These capabilities have been the subject of extensive research in recent years, and their application has been shown to significantly enhance the performance of LLMs across a variety of tasks. CoT prompting is a technique where language models are instructed to solve a task by following a sequence of reasoning steps, rather than providing an answer in a single response \cite{wei_chain--thought_2023}. Instructing language models in this way effectively allows them to perform symbolic operations, much like humans perform arithmetic operations by keeping track of intermediate steps.

The ability to use tools is another significant capability of language models \cite{schick_toolformer_2023}. This allows them to invoke external computational tools, thereby enriching their knowledge through querying search engines, accessing calculators, and so on \cite{schick_toolformer_2023}. These capabilities have been demonstrated to enhance the performance of large language models in a range of tasks that were previously inaccessible.
The recent advancements and results in revealing and exploiting the capabilities of LLMs suggest the potential for combining some of these capabilities to create more powerful and useful possibilities. This concept has been recently explored, leading to the development of the Modular Reasoning, Knowledge and Language (MRKL \cite{karpas_mrkl_2022}) and Reason+Act (ReAct \cite{yao_react_2023}) systems, which combine the CoT and tool-using capabilities of modern LLMs. By incorporating external tools into a CoT setting, agents of this type have recently been shown to outperform other methods based on large language models.

One direct benefit of effective tool usage is that it partially overcomes the unimodality issue of LLMs. Under this setting, they become capable of processing different types of input data, making real-time decisions in simulated environments, and even interacting with real-world robotic platforms. The solutions provided by LLMs to tasks also become more grounded in reality, as access to certain tools provides them with real, up-to-date information relevant to the task. This can, to some extent, limit the tendency of these models to generate unrealistic or “hallucinated” responses.

\subsubsection{Agents in chemistry: Unleashing the power from tools}
Despite their strengths as text generators and task solvers, and their remarkable few-shot and zero-shot performance, these models are also well known for their high propensity to generate false and inaccurate content, an issue that extends to easily verifiable matters such as basic arithmetic \cite{schick_toolformer_2023} and chemical operations \cite{dwhite_assessment_2023}. These limitations make the direct application of LLMs to chemistry a challenging matter.

The potential applications of large language models in chemistry were first explored in a large-scale collaboration involving researchers from around the world, an effort that resulted in the demonstration of 14 use-cases \cite{jablonka_14_2023}. The applications range from wrappers for computational tools, which enhance their accessibility by allowing natural language inputs to modify behaviors, to assistants for reaction optimization, and knowledge parsers and synthesizers for scientific question answering, among others. These are just a few of the possibilities that LLMs offer in chemistry, which, when combined with existing chemistry tools and databases, significantly increase the applicability and accessibility of computational applications.

More recently, Bran and Cox et al.\cite{bran_chemcrow_2023} extended the concept of LLM-powered agents for chemistry by curating and compiling a set of computational chemistry tools. Their system, ChemCrow, has been shown to be capable of planning and executing tasks in chemistry, effectively streamlining the reasoning process for several common chemical tasks across areas such as drug and materials design and synthesis. The authors demonstrate that this approach has a highly positive effect on LLM’s performance for tasks in chemistry, overcoming hallucinations and grounding their responses with data from reliable sources. Complementary approaches exist, with a sharper focus on cloud lab operability \cite{boiko_emergent_2023}.

The power of platforms like ChemCrow extends beyond merely serving as independent task solvers. They can be viewed as general chemistry assistants with the ultimate goal of making computational tools more accessible to chemists, thereby accelerating discovery. An additional highlight is the seamless exploitation of tool composability that this allows. It makes it straightforward to enrich the results of one tool with another, or to construct custom tool pipelines, all through simple requests in natural language.

\section{Outlook and final remarks}
Advancements in neural translation models, and specially with the introduction of the Transformer architecture, has sparked a revolution in machine learning for applications in chemistry and drug development. Analogies between chemical and natural language, and the publication of open databases and benchmarks, have inspired the representation of chemical tasks in the form of text, allowing straightforward application of Transformers to problems in this field.
This revolution has occurred in three stages, differentiated by the specificity of tasks. In a first stage, characterized by task-specificity and use of single-modality models, applications spanned molecule-to-molecule conversion tasks, like reaction outcome prediction and retrosynthetic planning, along with representation learning and downstream tasks like regression and classification. Their excellent performance and relative simplicity made them de-facto models in an array of applications.

In a second stage, researchers attempted to connect multiple additional modalities relevant to chemistry, like spectra from experiments, sequences of experimental actions, and even natural language, opening the way for an expanded number of applications involving modalities of any sort, however still task-specific. More recently, powered by vertiginous advancements in training and tuning of large language models, a series of works have been published that leverage a number of capabilities from such models. Among others, these contributions showcase applications in regression, classification, molecular generation and reaction optimization, all with unprecedented flexibility and usually improved performance over other methods. 
Another direction explores the integration of virtually unlimited modalities —in the form of tools— into agents powered by LLMs. The power of these agents has been demonstrated through a number of diverse tasks, ranging from molecular generation to automated organic synthesis, in an open-ended, highly customizable fashion. 

By leveraging the expressivity and flexibility of natural language, this last wave of applications aims to bridge the gap between the chemical and natural languages. As we continue to explore and harness these capabilities, we can look forward to a future where machine learning plays an even more integral role in accelerating scientific discovery.

\section*{Acknowledgements}

This work was created as part of NCCR Catalysis (grant number 180544), a National Centre of Competence in Research funded by the Swiss National Science Foundation.

\bibliography{sn-bibliography}

\end{document}